\documentclass{article}

\usepackage{iclr2022_conference,times}

\usepackage[utf8]{inputenc} 
\usepackage[T1]{fontenc}    
\usepackage{hyperref}       
\usepackage{url}            
\usepackage{booktabs}       
\usepackage{amsfonts}       
\usepackage{nicefrac}       
\usepackage{microtype}      
\usepackage[hyphenbreaks]{breakurl}

\usepackage{subcaption}

\usepackage{amsmath}
\usepackage{amssymb}
\usepackage{amsthm}

\newcommand{\ie}{i.e.~}
\newcommand{\eg}{e.g.~}
\newcommand{\wrt}{w.r.t.~}

\usepackage{enumitem}

\title{A Survey on Uncertainty Toolkits\\ for Deep Learning}

\author{Maximilian Pintz \\
University of Bonn, Fraunhofer IAIS\\
Schloss Birlinghoven\\
53757 Sankt Augustin, Germany \\
\footnotesize{\texttt{\{maximilian.alexander.pintz\}@iais.fraunhofer.de}} \hfill \\
\And
Joachim Sicking, Maximilian Poretschkin, Maram Akila \\
Fraunhofer IAIS\\
Schloss Birlinghoven\\
53757 Sankt Augustin, Germany \\
\footnotesize{\texttt{\{joachim.sicking,maximilian.poretschkin,maram.akila\}@iais.fraunhofer.de}} 
}

\iclrfinalcopy

\begin{document}

\maketitle
\begin{abstract} 
The success of deep learning (DL) fostered the creation of unifying frameworks such as tensorflow or pytorch as much as it was driven by their creation in return. Having common building blocks facilitates the exchange of, e.g., models or concepts and makes developments easier replicable.
Nonetheless, robust and reliable evaluation and assessment of DL models has often proven challenging.
This is at odds with their increasing safety relevance, which recently culminated in the field of ``trustworthy ML''.
We believe that, among others, further unification of evaluation and safeguarding methodologies in terms of toolkits, \ie small and specialized framework derivatives, might positively impact problems of trustworthiness as well as reproducibility. 
To this end, we present the first
survey on toolkits for uncertainty estimation (UE) 
in DL, as UE
forms a cornerstone in assessing model reliability. 
We investigate $11$ toolkits with respect to modeling and evaluation capabilities, providing an in-depth comparison for the three most promising ones, namely \textit{Pyro}, \textit{Tensorflow Probability}, and \textit{Uncertainty Quantification 360}.
While the first two provide a large degree of flexibility and seamless integration into their respective framework, the last one has the larger methodological scope.
\end{abstract}

\section{Introduction} 

Supervised deep learning (DL) has spurred 
progress in various application domains including computer vision \citep{Mahony2019DeepLV,chai_dlcv_2021}, natural language processing \citep{torfi_nlp_2020}, 
and speech recognition \citep{ALAM2020302} and is increasingly employed in safety-critical systems such as autonomous vehicles \citep{yurtsever_adsurvey_2020} or medical diagnosis systems \citep{hafiz_medical_2019}.
These systems potentially harm the environment, destroy equipment or put humans at risk \citep{safety-critical}, for instance, when vulnerable road users (in case of autonomous driving) or a severe disease (in case of a diagnostics system) are not recognized. 
The field of trustworthy ML \citep{Brundage2020TowardTA,chatila2021trustworthy,liu2021trustworthy,houben2021inspect} seeks to identify and mitigate such risks of ML systems to enable their responsible and safe operation. 
To this end, a more holistic notion of quality is proposed,
that extends beyond task performance and considers aspects like fairness, interpretability and reliability.
Central to the latter is the identification and handling of uncertainties, \ie factors that affect the system but are often not or only poorly accounted for.
Such uncertainties are ubiquitous in DL and stem, for instance, from data (\eg measurement noise and incorrect annotations) or variabilities in training procedures (\eg hyperparameters, initializations and limited training data).
Quantifying their impact on a system can contribute to its safety: for instance, an autonomous vehicle, that recognizes abnormal sensor input, may switch into a conservative travel mode or may activate redundant safety systems. \\[1em] 
The desirable adaptation of uncertainty estimation (UE) for a wide range of use cases \citep{sicking2022tailored} is often hindered by high implementation complexity. This adaptation could be accelerated by open-source toolkits,
which reduce this effort by providing easy-to-use building blocks and guidance in the form of minimal working examples. 
Being used and reviewed by the community, toolkits may moreover strengthen technical quality of UE components and advance common practices. 
The latter, \ie the use of common tools and standards, may drive further progress in the field as it facilitates easier reviews and reusing existing work. \\
In light of this, we survey deep uncertainty toolkits, to the best of our knowledge for the first time. 
The survey is structured as follows:
we first give a brief overview on DL software and UE (section \ref{sec:related_work}).
Next, we select $11$ publicly available open-source deep uncertainty toolkits from $180$ considered repositories (section \ref{sec:selection}) and present a catalog of criteria for their evaluation (section \ref{sec:criteria}).
We then apply these criteria to the collected toolkits in an comparative analysis (section \ref{sec:analysis}), describing their strengths and weaknesses. 
Finally, we highlight limitations and desirable future advancements of current deep uncertainty toolkits in the discussion (section \ref{sec:discussion}).

\section{Related work}\label{sec:related_work}

We first describe two types of DL software, DL frameworks and specialized toolkits that are discussed throughout this survey. Next, we give an overview over surveys on DL software and sketch their methodology and focus points in the second paragraph.
To provide a basis for the subsequent survey on UE toolkits, some foundational concepts of UE, especially \wrt their modeling and assessment, are reviewed in the third paragraph. 
Some simple UE capabilities are shipped as part of standard DL frameworks like tensorflow and pytorch. We outline these ``default'' capabilities and sketch their limitations in the last paragraph of the section.

\paragraph{DL frameworks and specialized toolkits}

There is a wide range of DL software covering different facets of DL. Two special types, namely DL frameworks and specialized toolkits, are in focus throughout this survey and described in the following. \\
\textit{DL frameworks} provide a comprehensive, reliable (well-tested/-maintained) and highly usable (modular structure, high-level interfaces) set of base components for building and evaluating deep learning systems. 
They typically provide layer modules that can be put together to form custom model architectures and gradient-descent-based optimizers for model training.
They are designed in respect of DL-specific software requirements including flexible data pipelines, automatic differentiation and efficient processing (GPU/multicore utilization, device distribution). Examples include tensorflow \citep{tensorflow2015-whitepaper}, pytorch \citep{pytorch} and MXNet \citep{chen_mxnet_2015}. \\
\textit{Specialized toolkits} extend the capabilities of DL frameworks.
They typically provide a collection of 
commonly used tools within a sub-field of deep learning. Examples include spaCy \citep{spacy} 
for natural language processing, 
GluonCV \citep{gluoncvnlp2020} for computer vision as well as the adversarial robustness toolbox \citep{nicolae_adversarial_2018}. The UE toolkits analyzed in this work fall into this category of DL software.

\paragraph{Surveys of DL software}\vspace*{-8pt}

Surveys on DL software concentrate mostly on the systematic assessment of DL frameworks \citep{nguyen_machine_2019,wang_survey_2019,landset_survey_2015,druzhov_survey_2016,Bahrampour2015ComparativeSO}.
Typical criteria for analysis include the range of provided components, GPU/multiprocessing support, integration of big data frameworks and general software criteria such as efficiency/scalability, ease of use and framework design/extensibility. 
Only few studies exist regarding specialized toolkits, such as \citet{zacharias_survey_2018} who concentrate on libraries for intelligent user interfaces or \citet{agarwal_interpretabletools_2020} who compare toolkits for interpretable machine learning under the aspects of range of provided methods and use cases. 
We seek to extend the overview on software for trustworthy AI with this survey on UE toolkits.

\paragraph{Elements of uncertainty estimation}\vspace*{-8pt}

In the following, we summarize central aspects of uncertainty estimation for the analysis of this survey. 
Further details to these aspects are provided in a methodological ``deep dive'' in the appendix (section \ref{sec:uncertainty_overview}). \\
DL systems are affected by several types of uncertainties \citep{huellermeier_aleatoric_2021}. 
A distinction can be made between reducible \textit{epistemic uncertainty} that arises from a lack of knowledge about the perfect model for solving a task, which includes lack of training data or approximation errors, and irreducible \textit{aleatoric uncertainty}, which includes inherent randomness in the data-generating process (\eg sensor noise or label ambiguity). \\ 
UE methods seek to quantify such uncertainties. 
In this survey we categorize them along two independent ``axes'': their \textit{integration depth} into a given model and the methodological framework they are based on. Along the former axis, we differentiate between (i) \textit{intrinsic methods} that integrate the uncertainty estimation directly into the architecture or training procedure (\eg dropout layers or ensemble training), (ii) \textit{post-hoc methods} that equip standard deep learning models with probability estimates and (iii) \textit{recalibration methods} that seek to improve existing uncertainty estimates \citep{guo_calibration_2017,kuhleshov_accurate_2018,Navrtil2021UncertaintyCC}. \\
The ``axis'' of \textit{methodological approaches} comprises the following. 
(i) Parametric likelihood methods consider networks that directly output distributional parameters instead of point estimates and are trained via likelihood optimization \citep{nix_estimating_1994,amini_deep_2020,sensoy_evidential_2018}. 
(ii) Bayesian neural networks (BNN) extend these approaches by incorporating (epistemic) uncertainty on network parameters. Examples include variational-inference based BNNs (VI-BNN, \citet{graves_practical_2011,blundell_weight_2015,rezende_variational_2015}), their variants based on dropout-sampling \citep{gal_dropout_2016,kendall_what_2017}),  Markov-chain Monte Carlo sampling (MCMC, \citet{neal_mcmc_2010,welling_bayesian_2011,li_preconditioned_2016}) or assumed density filtering (ADF-BNN, \citet{hernandez_lobato_probabilistic_2015,gast_lightweight_2018}). A related Bayesian approach are Gaussian processes (GP, \citet{rasmussen_gaussianprocess_2006}) and their scalable variants \citep{hensman_svgp_2013} that incorporate uncertainty in function space. 
(iii) Frequentist approaches directly leverage a combination of different models for UE, \eg deep ensembles \citep{lakshminarayanan_simple_2017,Wen2020BatchEnsembleAA,Durasov21} or the jackknife method \citep{giordano_swiss_2019,alaa_discriminative_2020,kim_predictive_2020}. \\
Several \textit{assessment techniques} have been developed that enable the benchmarking of uncertainty methods and the evaluation of their quality.  
Most commonly, these include (i) \textit{(proper) scoring rules} \citep{gneiting_proper_2007} that measure the fit of a predicted distribution to a ground-truth value (\eg negative log-likelihood (NLL) or brier score), (ii) \textit{calibration testing} \citep{guo_calibration_2017,kuhleshov_accurate_2018,Navrtil2021UncertaintyCC} that assesses alignment with a validation dataset (\eg reliability diagrams, excepted calibration error (ECE)), (iii) \textit{qualitative assessments} (\eg inspecting predicted distributions or the average predicted variance across a dataset) and (iv) the \textit{performance on auxiliary tasks}, \eg uncertainty-based separation of in-distribution and out-of-distribution datapoints as in \citet{ovadia_trust_2019}.

\paragraph{Uncertainty estimation in standard DL frameworks}\vspace*{-8pt}

Common DL frameworks natively provide basic building blocks for uncertainty estimation. 
TensorFlow/Keras\footnote{Keras was originally a separate library but has become a part of tensorflow. It provides high-level APIs for model building, training and evaluation.}, MXNet or Pytorch\footnote{The UE toolkits that we consider in our analysis build upon these three DL frameworks.} 
provide dropout functions that are originally intended for regularization \citep{srivastava2014dropout} but can also be used for building dropout-based uncertainty models.
Pytorch exhibits the {\small\texttt{distributions}} package, which implements backpropagation-compatible probability distributions using the reparametrization trick \citep{kingma_variational_2015}.
All mentioned frameworks support basic assessment techniques, \eg mean squared error or histograms. 
Despite these implementations, building a fully featured uncertainty model still typically requires significant implementation effort. 
For example, implementing a VI-BNN requires custom layer modules that impose a probability distribution on network parameters and the optimization of a custom objective function.
We instead focus on toolkits that build on these existing functionalities and provide a higher level of abstraction for reducing implementation effort.

\section{Selection of deep uncertainty toolkits} \label{sec:selection}

The first step of our systematic survey is to compile a collection of open-source toolkits on deep uncertainty estimation, which is the goal of this section.
For this, we first describe our procedure for 
compiling a pre-selection of potential code repositories on deep UE.
Next, we collect basic criteria a code repository needs to fulfill to be considered as an uncertainty toolkit in the scope of this work.
We finally apply these criteria to the pre-selection to obtain a collection of $11$ deep uncertainty toolkits that we use in our comparative analysis (see section \ref{sec:analysis}).

\paragraph{Search for code repositories related to UE in DL} \vspace*{-8pt}
In this survey, we focus on open-source code repositories that are available on {\small\verb|github.com|}.
We concentrate on github as it is the largest platform to host open-source code and to our knowledge hosts most of the publicly available deep uncertainty estimation software.
We use the public search function on github to find uncertainty-related code repositories. 
In particular, we make use of ``topics'', specific keywords for tagging repositories that allow other users to find these repositories more easily. 
In addition, users can ``star'' repositories they are especially interested in, which is a way of bookmarking the repository.
We examined the topics ``uncertainty-quantification'', ``uncertainty-estimation'' and also searched for the keywords ``uncertainty'' and ``probabilistic'' restricted to repositories of the topics ``machine-learning'' and ``deep-learning''.
In each case, we ordered the search results by the total number of ``stars'' they received so far and examined the first $30$ repositories in the resulting list, which adds up to a total of $180$ repositories. 
Among these repositories, there are deep uncertainty estimation libraries, various implementations accompanying research papers as well as software without a clear focus on deep learning, \eg general libraries on sensitivity analysis for numerical models.

\paragraph{Criteria constituting an uncertainty toolkit}\vspace*{-8pt}
As the focus of our work is on toolkits in supervised deep learning, we further filter the $180$ repositories. 
We count a software library as a \emph{deep uncertainty toolkit} if it contains multiple methods for at least one of the following:
\begin{itemize}[leftmargin=20pt,itemsep=0pt,topsep=-2pt,parsep=1pt]
    \item building and training a deep learning model that (inherently) supports UE (intrinsic methods),
    \item extending a given deep learning model with uncertainty estimates (post-hoc methods),
    \item improving already existing uncertainty estimates (recalibration) or
    \item assessing uncertainty estimates in terms of metrics or visualizations. 
\end{itemize}
Additionally, we require a toolkit to provide methods that are generally applicable to a broad range of scenarios, in particular to different datasets or network architectures. 
Sufficient documentation must be provided and either an application programming interface (API) or an (interactive) graphical user interface (GUI) application (in case of a standalone program).
The toolkit should have a clear focus on deep learning as evident from the documentation and should be integrated with common DL frameworks.

\paragraph{Selected toolkits}\vspace*{-8pt}

Out of the $180$ repositories, $11$ satisfy the above outlined criteria constituting uncertainty toolkits. They are listed in \autoref{tab:general_information_full} in the appendix. 
Among these, we find toolkits dedicated purely to deep uncertainty estimation (dedicated UE) as well as libraries with a broader scope that also provide UE capabilities.
The latter category includes \textit{GluonTS} (GTS, \citet{gluonts_jmlr}), a probabilistic time series library, and numerous deep probabilistic programming libraries (PPLs) that aim at specifying general Bayesian networks and performing inference for such models \citep{van2018introduction}. Namely these are \textit{Tensorflow Probability} (TFP, \citet{tfp}), \textit{Pyro} \citep{bingham2019pyro}, \textit{Edward2} (ED2, \citet{edward2}), \textit{ZhuSuan} (ZS, \citet{zhusuan2017}) and \textit{MXFusion} (MXF, \citet{mxfusion}). \\
Among the dedicated UE toolkits is, for instance, 
\textit{Uncertainty Quantification 360} (UQ360, \citet{uq360-june-2021}), which provides several uncertainty estimation tools and is part of a larger set of toolkits developed by IBM Research, each targeting individual dimensions of AI trustworthiness, including ``AI Fairness 360'' \citep{Bellamy2018AIF3} and ``AI Explainability 360'' \citep{arya_aiexplainability360_2020}. 
We also find several libraries with a more narrow range of functionality:
\textit{Uncertainty Toolbox} (UT, \citet{uncertainty_toolbox}) focuses on the recalibration and the assessment of uncertainty estimates in standard regression tasks.
\textit{Uncertainty Wizard} (UW, \citet{uncertainty_wizard}) is a package for extending keras models with dropout and ensemble-based uncertainty estimation capabilities. 
\textit{Bayesian Torch} (BT, \citet{bayesiantorch}) provides pytorch layer modules for building variational Bayesian neural networks and \textit{Keras-ADF} (KADF, \citet{keras-adf}) provides keras layer modules for assumed density filtering.
All of these toolkits are released under permissive free software licenses (namely Apache-2.0, MIT or BSD3) that come along with only minimal restrictions and allow, amongst others, their commercial usage.

\section{Evaluation criteria for uncertainty toolkits}\label{sec:criteria}

In this section, we detail our criteria that we use for analyzing the uncertainty estimation toolkits in section \ref{sec:analysis}. 
We divide both the criteria and the analysis in a ``core'' part, which we evaluate for all toolkits, and an extended more in-depth analysis for a selected subset of the most versatile tools, for which we also consider ``additional'' quality criteria. 
The ``core'' part focuses on the range of uncertainty modeling and evaluation techniques a framework supports as well as the neural architectures and data types/structures it is compatible with. 
The ``additional'' criteria moreover take code quality into account, both in terms of integration with standard DL frameworks as well as their level of documentation, modularity and testing.

\vspace{2mm}
\emph{Core criteria}

\paragraph{Range of supported uncertainty methods} \vspace*{-8pt}
Each uncertainty estimation method comes with its own set of strengths and weaknesses \wrt estimation quality, computational/storage costs, practicability (ease of implementation, training and evaluation), flexibility or theoretical soundness.
This implies that depending on the exact application scenario, some methods may be more suited than others.
Thus, we consider the range of functionalities for building or improving uncertainty models or extending standard (deterministic) models a relevant (core) evaluation criterion.

\paragraph{Range of supported evaluation techniques}\vspace*{-8pt}
A crucial part in research and development of DL models is the assessment of their prediction quality and the comparison against other models.
To enable such an assessment for uncertainty models, a toolkit should provide dedicated metrics for measuring the quality of uncertainty estimates, such as proper scoring rules, calibration or auxiliary scores.
As no metric is universally suited for every application scenario, a toolkit should cover a wide range of different uncertainty metrics, potentially aided by visualizations (\eg calibration plots or confidence bands). 

\paragraph{Range of supported architectures and data structures} \vspace*{-8pt}
A toolkit that only supports the multi-layer perceptron architecture hardly finds use in computer vision tasks due to its lack of support for convolutional layers or image inputs.
This illustrates that to be applicable to different use cases, a toolkit should support a wide range of network architectures, optimizers and data structures, which we consider a relevant criterion for our evaluation.

\vspace{2mm}
\emph{Additional criteria}

\paragraph{Integration with DL frameworks}\vspace*{-8pt}
Deep learning software in general comes with a special set of requirements that include flexible data pipelines, efficient optimization (automatic differentiation, GPU/multicore utilization), reproducibility (logging of hyperparameters, seeds and configurations) and modular model building.
DL frameworks like tensorflow or pytorch are (i) geared to these desiderata, (ii)  well-maintained and exhibit a large community of developers and users and (iii) can be seen as de-facto standards when developing deep learning software.
In order to profit from their properties, an uncertainty toolkit should actively employ tools from DL frameworks and exhibit high interoperability with them, \eg by providing layer modules that can be inserted into existing framework-native models.

\paragraph{Software quality} \vspace*{-8pt}
We also evaluate the toolkits in terms of general software quality criteria including (i) usability and documentation quality, (ii) modularity and integrability and (iii) maintenance and testing.
A highly usable toolkit is, among others, delivered with
a simple installation procedure, a code architecture/API that is easy to understand and to work with and is well-documented.
A highly modular toolkit provides flexible building blocks, ideally at different abstraction levels from high-level (being easy-to-use, but less flexible) to low-level (highest level of customization, but higher implementation effort and a higher degree of technical understanding required) that can be combined with each other. 
Important aspects of code maintenance include actively supporting code contributions, heeding coding style conventions (\eg via code linting) and enforcing code testing (\eg continuous integration and reporting code coverage).

\section{Comparative analysis of the selected uncertainty toolkits}\label{sec:analysis}

In the following, we first analyze all uncertainty toolkits from section \ref{sec:selection} \wrt the core criteria of supported uncertainty methods, evaluation techniques and architectures/data structures (cf.\ section \ref{sec:criteria}). 
Based on this analysis, we select the three most relevant toolkits, namely UQ360, Pyro and TFP, for an in-depth analysis in \autoref{sec:detailed_analysis}.
This will, additionally, include the extended set of evaluation criteria from \autoref{sec:criteria}. 
We structure the analysis in both cases along the criteria (and not along the toolkits) to better contrast the toolkits against one another. 

\subsection{Analysis with respect to core criteria}\label{sec:core_analysis}

We now briefly compare all $11$ uncertainty toolkits selected in \autoref{sec:selection} \wrt the core criteria. A concise summary of these findings per toolkit can be found in the upper part of \autoref{tab:analysis_summary_full} in the appendix.

\paragraph{Range of supported uncertainty methods}\vspace*{-8pt}
UQ360 provides the widest range of methods, covering intrinsic, post-hoc and recalibration methods. 
All other toolkits cover only intrinsic methods, except for UT, which has a narrow focus on basic recalibration methods.
The deep PPLs provide intrinsic Bayesian methods, mainly MCMC, VI-BNNs or Gaussian processes. In contrast to comparable methods from BT or UQ360, they typically cover a broader range of prior, likelihood and variational posterior distributions.
The time series library GTS only provides parametric likelihood methods, but supports a large range of distributions (especially compared to UQ360 that only supports Gaussian likelihoods).
UW, BT and KADF are dedicated UE libraries with a narrow focus on the intrinsic methods dropout/ensembling, VI-BNN and ADF-BNN, respectively.

\paragraph{Range of supported evaluation techniques}\vspace*{-8pt}
UQ360 and UT have the most comprehensive set of assessment tools ranging from scoring rules over calibration scores and plotting functions, followed by TFP, which lacks plotting functions.
GTS provides scoring rules and a function for plotting forecasted time series with confidence bands. 
In comparison, Pyro, ED2 and ZS have more narrow capabilities and focus on the standard evaluation metric in Bayesian inference, namely computing log-probabilities of the predictive distribution. 
Some toolkits (UW, BT, KADF and MXF) do not contain any assessment capabilities.

\paragraph{Range of supported architectures and data structures}\vspace*{-8pt}
Deep PPLs are constructed to enable Bayesian inference for a broad range of models with different architectures.
UW, BT and KDF provide support for classification
and regression models. The former provides a keras-based model interface for this purpose, while BT and KDF provide drop-in replacements for dense and convolutional layers.
BT additionally covers recurrent models by including probabilistic long short-term memory layers \citep{hochreiter1997long}.
In contrast, GTS and UT focus on specific domains such as time series modelling (GTS) or 1D regression (UT). 

Overall, we find TFP, Pyro and UQ360 to provide the most comprehensive catalog of models, assessment techniques and supported architectures.

\subsection{Detailed analysis of TFP, Pyro and UQ360}\label{sec:detailed_analysis}

We now extend our analysis for the three most promising toolkits, namely UQ360, Pyro and TFP.
In particular, we provide more details \wrt the core criteria and consider the two ``additional'' criteria, integration with DL frameworks and software quality.
For a tabular summary of the results, see the bottom part of \autoref{tab:analysis_summary_full} in the appendix.

\paragraph{Range of supported uncertainty methods}\vspace*{-8pt}
As discussed before, UQ360 provides a wide range of methods at different levels of integration depth.
Specifically, its intrinsic methods comprise pure aleatoric uncertainty estimation via Gaussian likelihoods as well as joint modeling approaches for aleatoric and epistemic uncertainty via VI-BNNs or deep ensembles.
Post-hoc methods include the infinitesimal jackknife \citep{giordano_swiss_2019} and surrogate model approaches \citep{chen2019confidence}.
UQ360 moreover is the only toolkit with recalibration methods other than standard Platt scaling or isotonic regression and additionally includes \eg auxiliary interval predictors \citep{thiagarajan_auxiliary_2020} and UCC rescaling \citep{Navrtil2021UncertaintyCC}. \\
The PPLs TFP and Pyro cover intrinsic methods such as parametric likelihood (aleatoric uncertainty), VI-BNNs (aleatoric + epistemic) and (variational) Gaussian processes (functional uncertainty).
They also provide non-parametric posterior sampling methods such as Hamiltonian Monte Carlo \citep{neal_mcmc_2010} or general Metropolis-Hastings \citep{metropolis,hastings}, which may be useful for researchers due to its high estimation accuracy, but is typically infeasible to apply in large-scale models.
A more scalable variant, SGLD \citep{welling_bayesian_2011,li_preconditioned_2016}, is additionally provided by TFP. 
By design, the PPLs support a much broader range of prior, likelihood and variational posterior distributions compared to the corresponding modules of UQ360. For example, it is possible to employ normalizing flows \citep{rezende_variational_2015} for more flexible priors or posteriors.

\paragraph{Range of supported evaluation techniques} \vspace*{-8pt}
UQ360 offers a wide range of assessment techniques. 
It covers standard metrics, namely the Gaussian NLL for regression, the classification ECE and the Brier score as well as less frequently used scores such as the area under the risk-rejection-rate curve \citep{vojtech_discriminative_2019} or the uncertainty calibration curve (UCC, \citet{Navrtil2021UncertaintyCC}).  
Apart from these metrics, it has helpful ``tools'' for assessment such as a function that decomposes samples of class probabilities into aleatoric and epistemic uncertainty or various plotting functions (\eg calibration curves and prediction intervals). \\
In comparison, TFP has more narrow assessment capabilities when it comes to comparing uncertainty estimates with ground truth labels and provides the above mentioned standard metrics for this purpose.
Instead of scalar values, a TFP model predicts distribution objects. These provide functions for computing standard statistics such as log-probabilities, moments, quantiles, correlation, cumulative distribution functions (cdf's) and KL divergences to other distributions. 
These statistics can be computed for a wide range of distributions besides the Gaussian, including \eg multivariate distributions and mixture models. \\
Pyro offers a module for approximate sampling from the predictive distribution of a given Bayesian model, which serves as a basis for further assessment.
For this, it provides general statistical utilities (\eg for computing quantiles, autocorrelation or prediction intervals) and the CRPS \citep{gneiting_proper_2007} as a scoring rule.

\paragraph{Range of supported architectures and data structures} \vspace*{-8pt}
UQ360's method catalog covers regression and classification scenarios (mostly in 1D). 
However, many model classes (\eg VI-BNN or deep ensembles) have pre-written training procedures or even architectures, which limits the ability to extend a given deep learning model with uncertainty estimates. 
Inputs are typically required to be of (2D) tabular form, limiting the ability to deal with sequential data or other data types.
In contrast, TFP is a collection of smaller building blocks and provides layer and optimizer modules that can be used in addition to or as drop-in replacements for the standard tensorflow modules. 
It provides, for instance, probabilistic drop-in replacement modules for dense or convolutional layers to turn given NNs into BNNs.
For pytorch, Pyro provides comparable capacities for turning NNs into BNNs, essentially by subclassing the respective NN classes and replacing the pytorch parameters with stochastic modules. 
The highly modular structure of both PPLs allows to build models with a wide variety of different architectures and inputs, for instance, for multidimensional regression and for sequential data.

\paragraph{Integration with deep learning frameworks} \vspace*{-8pt}
As Tensorflow Probability (TFP) is a part of the tensorflow ecosystem, its modules are designed to interface seamlessly with other modules of tensorflow/keras.
In particular, it integrates with the keras model API, which allows the developer to build and train probabilistic models using keras' native {\small\verb|compile|} and {\small\verb|fit|} functions.
Pyro utilizes layer, optimizer, data and distribution functions from its base framework pytorch.
When building BNNs in Pyro, one can use standard pytorch modules (\eg layers and activation functions) in addition to the newly provided stochastic modules. 
Moreover, models and inference procedures can be encapsulated as pytorch modules, which enables the creation of TorchScript programs for deployment. \\
Most model classes of UQ360 are based on pytorch. These are easy-to-use, but not as flexible and modular as TFP's building blocks for keras models. Changing network architectures, input pipelines or the training procedure often requires modifications to the code, instead of simply passing different modules (\eg optimizers and dataloaders) to the model class. 

\paragraph{Software quality} \vspace*{-8pt}
All three toolkits have a simple installation procedure and are listed on the Python package index. 
The toolkits and all their dependencies can be installed via a single pip command, except for TFP, which requires installing a compatible version of tensorflow beforehand. 
Pyro additionally provides, besides pip, a docker image for installation.
There is a sensible choice of dependencies across the toolkits, using a base DL framework plus standard libraries like numpy, scipy, matplotlib or pandas.
All toolkits are actively maintained, welcome to post issues and pull requests and provide an explicit documentation for contributing (via a readme).
TFP and Pyro follow coding style guides (e.g. PEP8 \footnote{PEP8 are coding style guidelines used by the official python distribution.}) to ensure that newly added code fits to coding conventions and, moreover, employ continuous integration for code testing. \\
We next discuss interface structures of the toolkits.
UQ360 has a simple scikit-learn-esque API (\ie model classes providing {\small\verb|fit|} and {\small\verb|predict|} functions taking numpy arrays as input) that is equally intuitive to understand. 
TFP employs the keras model API, which is highly flexible and also easy to understand. 
Pyro's API puts more emphasis on the Bayesian modelling perspective of uncertainty estimation. 
The user needs to define two separate classes (or functions), where one defines the model with its prior and likelihood distributions and the other (the so-called guide) defines the variational posterior distribution. 
As Pyro uses an API structure different from the widely known APIs (for instance of such as sklearn or keras), it might be more difficult for users to quickly get started with it.
On the other hand, the possibility to automatically generate standard posterior distributions (e.g. mean-field VI) or to create constrained network parameters (e.g. positive scale parameters for distributions) improves API usability. In comparison to TFP, Pyro misses direct drop-in replacements for standard layers, which results in a higher implementation effort for customizations. \\
The code of UQ360 is simple and highly readable, but could more frequently employ input checks (\eg for types or shapes) and proper exception handling.
The code bases of Pyro and TFP are designed to be flexible and to satisfy compatibility requirements (\eg implementing functions of superclasses) of their respective base library (pytorch or tensorflow) and have (partly as a result from this) a more complex and covert structure. 
It is of high overall quality, which manifests in appropriate input checking and exception handling and in a clean and consistent coding style. 

\section{Discussion}\label{sec:discussion} 

Reliable models should be able to identify the boundaries at which they function properly. 
Finding sources of uncertainty and quantifying their impact on the model performance contributes to this aspect.
While many approaches to uncertainty estimation have been developed, there are entry barriers on their use, including high technical complexity.
By providing high-quality software components, toolkits for UE help to overcome such entry barriers and additionally facilitate standardized evaluation.
To help the reader in selecting an appropriate toolkit, we provide the first survey on existing deep uncertainty toolkits.
We defined minimum requirements for such toolkits and analyzed $11$ of them with respect to range of supported uncertainty methods, evaluation techniques, architectures and data structures.  
The three most relevant ones (TFP, Pyro and UQ360) were additionally examined under the aspects of integration with DL frameworks and software quality.
All analyzed toolkits provide modules to ease the development and assessment of uncertainty models. 
They encompass deep probabilistic programming libraries (\eg Pyro, MXF, ZS) that focus on infusing Bayesian inference into DL models as well as toolkits dedicated to UE (\eg UQ360, UT) that cover a broader range of methods or assessments.
We plan to extend our detailed analysis to more toolkits in future iterations of this work.
The survey also reveals desirable improvements to UE software and future incentives that are discussed in the following.

\paragraph{Extend the technical capabilities of uncertainty toolkits}\vspace*{-8pt} 
The considered UE toolkits are either more comprehensive (\eg UQ360) or more modular and interoperable with DL frameworks (\eg TFP, Pyro, UW).
An UE toolkit should ideally combine both aspects.
Further concrete means for improving the technical comprehensiveness of currently available toolkits include: (i) providing more post-hoc methods, which are of high interest in scenarios where rebuilding a model or retraining is associated with high costs; (ii) infusing uncertainty into application-specific network components, \eg into non-maximum suppression or clustering procedures of computer vision models (as in \citet{meyer_lasernet_2020,harakeh_bayesod_2020}); (iii) providing tools for task performance benchmarking that account for uncertainties (\eg based on hypothesis testing as in \citep{gorman-bedrick-2019-need}).  
\paragraph{Interfacing with other toolkits}\vspace*{-8pt} 

There is a multitude of relevant tasks and problems in deep learning besides uncertainty estimation.
For example, different metrics and approaches are considered to assess and ensure different aspects of AI trustworthiness, such as robustness, interpretability or fairness.
To address multiple such problems simultaneously employed toolkits should provide interoperable interfaces.
This might be facilitated if, as we evaluated, the toolkits integrate seamlessly with the (same) underlying DL framework.
Such interoperability between toolkits can even directly impact UE.
For instance, data augmentation toolkits (\eg augLy \citep{papakipos2022augly}) provide input corruptions for out-of-distribution assessments and  online-learning toolkits (\eg river \citep{2020river}) can be used to adapt uncertainty models to future observations.
A future prospect is the combination of several toolkits on trustworthy AI (\eg AI Fairness 360 \citep{Bellamy2018AIF3}, Adversarial Robustness Toolbox \citep{nicolae_adversarial_2018}) into a highly comprehensive testing framework.

\paragraph{Guidance on correct tool usage}\vspace*{-8pt}
By providing usable high-level interfaces, toolkits reduce an entry barrier against employing (potentially highly complex) algorithms by a broad range of users.
They should, additionally, support correct usage of the provided tools, \eg by providing comprehensive guidelines.
UQ360's documentation provides guidance on choosing an uncertainty method and on communicating uncertainty estimates to stakeholders, which we see as a step in the right direction.
Additional means that support a more informed and effective use of UE include the attribution of uncertainty to concrete sources (\ie how much uncertainty arises from data, hyperparameter tuning, initializations, etc.), instructions on reducing uncertainties as well as contrasting evaluation scores against one another and describing their properties. 
Visual and interactive interfaces can aid correct tool use further.
To generally improve documentation practices, standards for documenting toolkits and their tools could be developed, compare for instance ``model cards'' \citep{mitchell_modelcard_2019}.

\paragraph{Maintaining code quality}\vspace*{-8pt}
Open-source seems to be a prerequisite for trustworthy implementations of evaluation standards and many of the reviewed toolkits principally place value on code quality as evident by supporting code contributions or employing automated testing tools.
But, this is not always sufficient to ensure high quality and absence of critical bugs. 
Especially in the context of Trustworthy AI and safety-critical systems, which we see as a large application field for UE, this can become an issue.
To further improve in that regard, it seems desirable to incentivize systematic code reviews, \eg by introducing bug bounty programs for major toolkits and generally calling greater attention towards this topic (\eg via dedicated workshops on code quality). 

\subsubsection*{Acknowledgments}

The development of this publication was supported by the Ministry of Economic Affairs, Innovation, Digitalization and Energy of the State of North Rhine-Westphalia as part of the flagship project ZERTIFIZIERTE KI.

\bibliography{references}
\bibliographystyle{iclr2022_conference}

\newpage
\section*{Appendix}

\subsection{Detailed overview on uncertainty modeling and assessment} \label{sec:uncertainty_overview}

The following section gives an literature overview over uncertainty estimation in DL, describing sources of uncertainty, approaches and applications for uncertainty estimation and assessment. 
It provides further details to the paragraph on ``elements of uncertainty estimation'' in \autoref{sec:related_work}. 

Neural networks are subject to several kinds of uncertainties \citep{huellermeier_aleatoric_2021}. 
We generally do not know whether the chosen neural network model is optimal for the given task. There are also approximation errors caused by the optimization procedure (\eg due to (suboptimal) hyperparameter choices, random initializations, lack of data).
Such uncertainties that arise from lack of knowledge about the optimal model for a given task are subsumed under the term \textit{epistemic uncertainty}, which is theoretically reducible \eg by selecting better models or supplying more training data.
Additionally, there is irreducible \textit{aleatoric uncertainty}, which is stochasticity in the outcome of the training procedure caused by inherently random factors, such as most importantly the data (\eg sensor noise, mislabeled inputs or imperfect information).
Besides for quantifying prediction and estimation variance, uncertainty estimation has also been explored for auxiliary tasks, such as detecting out-of-distribution inputs \citep{ovadia_trust_2019,vyas_ood_2018}, active learning \citep{gal_deep_2017,beluch_power_2018}, detection of adversarial examples \citep{ritter_scalable_2018,amini_deep_2020} and continual learning \citep{osawa2019practical,nguyen2018variational}. 
The main tool for deep uncertainty estimation are probability distributions, typically over network outputs. The expected value serves as main network prediction and uncertainty is quantified in terms of variance, quantiles or entropy.
Approaches for uncertainty estimation include the following (see \citet{gawlikowski_survey_2021} for a comprehensive survey):
\begin{itemize}[leftmargin=20pt,itemsep=0pt,topsep=-2pt,parsep=1pt]
    \item In parametric likelihood (PL) methods, the network outputs distributional parameters instead of point estimates and is trained via likelihood optimization \citep{nix_estimating_1994,amini_deep_2020,sensoy_evidential_2018}. They are commonly used to estimate aleatoric uncertainty and are combined with other methods to incorporate epistemic uncertainty in addition \citep{kendall_what_2017}.
    \item Bayesian neural networks (BNNs) extend the parametric likelihood approach and incorporate (epistemic) uncertainty on network parameters by estimating a posterior and predictive distribution based on Bayes rule. Different posterior estimation procedures have been considered for DL, including variational inference (VI-BNNs, \citet{graves_practical_2011,blundell_weight_2015,rezende_variational_2015}) with dropout sampling \citep{gal_dropout_2016,kingma_variational_2015} being an important variant, expectation propagation \citep{hernandez_lobato_probabilistic_2015,gast_lightweight_2018}, Markov-chain Monte Carlo sampling (MCMC, \citet{neal_mcmc_2010,welling_bayesian_2011,li_preconditioned_2016}) or the Laplace approximation \citep{mackay_practical_1992,ritter_scalable_2018}.
    Gaussian processes (GP, \citet{rasmussen_gaussianprocess_2006}) are a related Bayesian approach that incorporate uncertainty in a more general function space, instead of the network's parameter space. Standard GPs are computationally infeasible for large-scale models. However, there are scalable variants \citep{hensman_svgp_2013} that can even be used as a differentiable network component \citep{tran_variational_2016} and as a building block for UE in deep learning models \citep{liu2020simple}.
    \item Frequentist approaches directly leverage a combination of different models for uncertainty estimation.
        Common approaches are based on ensembling \citep{lakshminarayanan_simple_2017,Wen2020BatchEnsembleAA,Durasov21}) or the jackknife method \citep{giordano_swiss_2019,alaa_discriminative_2020,kim_predictive_2020}.
\end{itemize}
Uncertainty methods can be further categorized into (i) intrinsic methods, that integrate the uncertainty estimation directly into the architecture or training procedure (\eg ensemble training \citep{lakshminarayanan_simple_2017} and dropout-based approaches \citep{gal_dropout_2016,kendall_what_2017,fan2021contextual,sicking2021wdropout}), (ii) post-hoc methods, that extend standard deep learning models (\eg Laplace approximation, infinitesimal jackknife, surrogate models \citep{chen2019confidence}) and (iii) recalibration methods, that seek to improve existing uncertainty estimates \citep{guo_calibration_2017,kuhleshov_accurate_2018,Navrtil2021UncertaintyCC}. \\
We now discuss several techniques for assessing the quality of uncertainty estimates.
\begin{itemize}[leftmargin=20pt,itemsep=0pt,topsep=-2pt,parsep=1pt]
    \item (Proper) Scoring rules are point-wise metrics that measure for the fit of a predicted distribution to a ground-truth value and are typically evaluated on held-out validation datasets. Common scoring rules include the brier score, negative log-likelihood (NLL), continuous ranked probability score (CRPS) or the interval score (see \citet{gneiting_proper_2007}).
\item Confidence calibration measures the alignment of confidence estimates with a validation dataset (\eg a $90\%$ confidence interval should contain $90\%$ correct labels).
Calibration can be visually assessed via reliability diagrams or curves \citep{guo_calibration_2017,Navrtil2021UncertaintyCC} that plot confidence levels against a correctness metric (\eg classification accuracy).
Quantitative metrics typically consider the area under the aforementioned curves, most notably the expected calibration error (ECE) \citep{guo_calibration_2017,kuhleshov_accurate_2018}.
Adversarial group calibration \citep{zhao2020individual} is a more strict variant of calibration assessment that considers alignment of confidence estimates with every subset of the dataset, instead of only the whole dataset. 
\item Uncertainty estimates can also be qualitatively assessed, \eg by visually inspecting predicted distributions, confidence intervals against ground-truth values. Considering the average uncertainty across a dataset (also called sharpness) is crucial in calibration assessments as argued by \citet{kuhleshov_accurate_2018}.
\item Auxiliary scores measure the performance of uncertainty estimates in auxiliary tasks such as out-of-distribution detection.
Standard evaluation methods for binary threshold classifiers (\eg separating true from false detections) include precision-recall and receiver operating curves for visual assessment or computing average precision or AUCROC scores.
Another approach is to compare histograms of uncertainty estimates (\eg variance or predictive entropy) on inputs from each class (\eg in-distribution or out-of-distribution), as done in \citet{ovadia_trust_2019}. 
\end{itemize}

\subsection{Comparative analysis of the selected uncertainty toolkits}

This section provides two tables with additional information about the toolkits. 
\autoref{tab:general_information_full} lists all toolkits with their license, intended purpose and the exact version that has been considered in this survey. 
\autoref{tab:analysis_summary_full} provides a concise summary of the results of our analysis from \autoref{sec:analysis}.

\begin{table}[ht]
  \caption{General information on the examined uncertainty estimation toolkits. All toolkits use python as programming language. We provide the github commit number in cases where no version number is given.  
  }
  \label{tab:general_information_full}
  \centering
  \begin{small}
  \begin{tabular}{lp{30mm}lp{15mm}p{20mm}p{20mm}}
    \toprule
    \textbf{UE toolkit} & \textbf{Developer} & \textbf{License} & \textbf{Version/ commit} & \textbf{Base libraries/ frameworks} & \textbf{Toolkit type}  \\
    \midrule
    TFP & \citet{tfp} (Google Brain) & Apache-2.0 & {\small\verb|0.15.0|} & Tensorflow/ Keras & PPL \\ 
    Pyro & \citet{bingham2019pyro} (Uber AI Labs) & Apache-2.0 & {\small\verb|1.8.0|} & Pytorch, JAX & PPL \\
    MXF & \citet{mxfusion} (Amazon Web Services) & Apache-2.0 & {\small\verb|0.3.1|} & MXNet & PPL \\
    ZS & \citet{zhusuan2017} & MIT & {\small\verb|4386b2a|} & Tensorflow & PPL \\
    ED2 & \citet{edward2} (Google Brain) & Apache-2.0 & {\small\verb|f420d83|} & Tensorflow/ Keras  & PPL \\ 
    GTS & \citet{gluonts_jmlr} (Amazon Web Services) & Apache-2.0 & {\small\verb|0.9.0|} & MXNet, Pytorch & Time series \\ 
    UQ360 & \citet{uq360-june-2021} (IBM Research)  & Apache-2.0 & {\small{\verb|2378bfa|}} & Pytorch & Dedicated UE  \\ 
    UT & \citet{uncertainty_toolbox} & MIT & {\small\verb|v0.1.0|}  & Scipy, Scikit-learn & Dedicated UE \\
    UW & \citet{uncertainty_wizard}  & MIT & {\small\verb|v0.2.0|} & Tensorflow/ Keras & Dedicated UE \\
    BT & \citet{bayesiantorch} (Intel Labs) & BSD3 & {\small\verb|99876f3|} & Pytorch & Dedicated UE \\
    KADF & \citet{keras-adf} & MIT & {\small{\verb|19.1.0|}} & Tensorflow/ Keras & Dedicated UE \\ 
    \bottomrule
  \end{tabular}
  \end{small}
\end{table}

\begin{table}[ht]
  \caption{Analysis summary of the uncertainty toolkits selected in \autoref{sec:selection}. The top table considers all toolkits with respect to our core criteria (cf. \autoref{sec:criteria}). The bottom table provides additional criteria and considers the three best-performing toolkits with respect to the core criteria. ``Standard statistics'' refers to general (sample) measures such as variance, quantiles or correlation.}
  \label{tab:analysis_summary_full}
  \centering
  \begin{scriptsize}
  \begin{tabular}{lp{38mm}p{38mm}p{38mm}}
    \toprule
    \textbf{UE toolkit} & \textbf{Range of supported uncertainty methods} & \textbf{Range of supported evaluation techniques} & \textbf{Range of supported architectures and data structures} \\
    \midrule
    TFP & 
    \begin{minipage}[t]{\linewidth}
    \begin{itemize}[leftmargin=5pt,itemsep=0pt,topsep=0pt,parsep=0pt] 
    \item intrinsic (PL, VI-BNN, GP, MCMC)
    \item high range of supported distributions
    \end{itemize}
    \end{minipage} & 
    \begin{minipage}[t]{\linewidth}
    \begin{itemize}[leftmargin=5pt,itemsep=0pt,topsep=0pt,parsep=0pt]
    \item broad range of standard statistics
    \item scoring rules (NLL, Brier)
    \item classification calibration (ECE)
    \end{itemize}
    \end{minipage} & 
    \begin{minipage}[t]{\linewidth}
    \begin{itemize}[leftmargin=5pt,itemsep=0pt,topsep=0pt,parsep=0pt]
    \item regression, classification, sequential
    \end{itemize}
    \end{minipage} \\[6ex]
    Pyro & 
    \begin{minipage}[t]{\linewidth}
    \begin{itemize}[leftmargin=5pt,itemsep=0pt,topsep=0pt,parsep=0pt]
    \item intrinsic (PL, VI-BNN, GP, MCMC)
    \item high range of supported distributions
    \end{itemize}
    \end{minipage} & 
    \begin{minipage}[t]{\linewidth}
    \begin{itemize}[leftmargin=5pt,itemsep=0pt,topsep=0pt,parsep=0pt]
    \item standard statistics
    \item scoring rules (NLL, CRPS)
    \end{itemize}
    \end{minipage} & 
    \begin{minipage}[t]{\linewidth}
    \begin{itemize}[leftmargin=5pt,itemsep=0pt,topsep=-8pt,parsep=0pt] 
    \item regression, classification, sequential
    \end{itemize}
    \end{minipage} \\[6ex]
    ED2 & 
    \begin{minipage}[t]{\linewidth}
    \begin{itemize}[leftmargin=5pt,itemsep=0pt,topsep=0pt,parsep=0pt]
    \item intrinsic (PL, VI-BNN, GP, MCMC)
    \item high range of supported distributions
    \end{itemize}
    \end{minipage} & 
    \begin{minipage}[t]{\linewidth}
    \begin{itemize}[leftmargin=5pt,itemsep=0pt,topsep=0pt,parsep=0pt]
    \item narrow range of scoring rules (NLL)
    \end{itemize}
    \end{minipage} & 
    \begin{minipage}[t]{\linewidth}
    \begin{itemize}[leftmargin=5pt,itemsep=0pt,topsep=0pt,parsep=0pt]
    \item regression, classification, sequential
    \end{itemize}
    \end{minipage} \\[6ex]
    ZS & 
    \begin{minipage}[t]{\linewidth}
    \begin{itemize}[leftmargin=5pt,itemsep=0pt,topsep=0pt,parsep=0pt]
    \item intrinsic (PL, VI-BNN, MCMC) 
    \item high range of supported distributions
    \end{itemize}
    \end{minipage} & 
    \begin{minipage}[t]{\linewidth}
    \begin{itemize}[leftmargin=5pt,itemsep=0pt,topsep=0pt,parsep=0pt]
    \item narrow range of standard statistics
    \item narrow range of scoring rules (NLL)
    \end{itemize}
    \end{minipage} & 
    \begin{minipage}[t]{\linewidth}
    \begin{itemize}[leftmargin=5pt,itemsep=0pt,topsep=0pt,parsep=0pt]
    \item regression, classification, sequential 
    \end{itemize}
    \end{minipage} \\[6ex]
    MXF & 
    \begin{minipage}[t]{\linewidth}
    \begin{itemize}[leftmargin=5pt,itemsep=0pt,topsep=0pt,parsep=0pt]
    \item intrinsic (PL, VI-BNN, GP) 
     \end{itemize}
    \end{minipage} & [none] & 
    \begin{minipage}[t]{\linewidth}
    \begin{itemize}[leftmargin=5pt,itemsep=0pt,topsep=0pt,parsep=0pt]
    \item regression, classification
    \end{itemize}
    \end{minipage} \\[6ex]
    GTS & 
    \begin{minipage}[t]{\linewidth}
    \begin{itemize}[leftmargin=5pt,itemsep=0pt,topsep=-8pt,parsep=0pt]
    \item intrinsic (PL) 
    \end{itemize}
    \end{minipage} & 
    \begin{minipage}[t]{\linewidth}
    \begin{itemize}[leftmargin=5pt,itemsep=0pt,topsep=-8pt,parsep=0pt]
    \item scoring rules
    \item plotting function for conf. intervals
    \end{itemize}
    \end{minipage} & 
    \begin{minipage}[t]{\linewidth}
    \begin{itemize}[leftmargin=5pt,itemsep=0pt,topsep=-8pt,parsep=0pt]
    \item sequential
    \end{itemize}
    \end{minipage} \\[6ex] 
    UQ360 & 
    \begin{minipage}[t]{\linewidth}
    \begin{itemize}[leftmargin=5pt,itemsep=0pt,topsep=0pt,parsep=0pt]
        \item intrinsic (PL, VI-BNN, ensembling)
        \item post-hoc (jackknife-based, surrogate models)
        \item recalibration
    \end{itemize}
    \end{minipage} & 
    \begin{minipage}[t]{\linewidth}
    \begin{itemize}[leftmargin=5pt,itemsep=0pt,topsep=0pt,parsep=0pt]
        \item scoring rules
        \item calibration assessment (ECE)
        \item plotting functions
    \end{itemize}
    \end{minipage} & 
    \begin{minipage}[t]{\linewidth}
    \begin{itemize}[leftmargin=5pt,itemsep=0pt,topsep=0pt,parsep=0pt]
        \item regression, classification
        \item tabular inputs
        \item low flexibility in intrinsic methods
    \end{itemize}
    \end{minipage} \\[12ex]
    UT & 
    \begin{minipage}[t]{\linewidth}
    \begin{itemize}[leftmargin=5pt,itemsep=0pt,topsep=0pt,parsep=0pt] 
    \item basic recalibration
    \end{itemize}
    \end{minipage} &  
    \begin{minipage}[t]{\linewidth}
    \begin{itemize}[leftmargin=5pt,itemsep=0pt,topsep=0pt,parsep=0pt] 
    \item broad range of scoring rules
    \item calibration assessment (ECE, AGC)
    \item plotting functions
    \end{itemize}
    \end{minipage} & 
    \begin{minipage}[t]{\linewidth}
    \begin{itemize}[leftmargin=5pt,itemsep=0pt,topsep=0pt,parsep=0pt] 
    \item 1D-regression
    \end{itemize}
    \end{minipage} \\[6ex]
    UW & 
    \begin{minipage}[t]{\linewidth}
    \begin{itemize}[leftmargin=5pt,itemsep=0pt,topsep=0pt,parsep=0pt] 
    \item intrinsic (ensembling, dropout) 
    \end{itemize}
    \end{minipage} &  [none] & 
    \begin{minipage}[t]{\linewidth}
    \begin{itemize}[leftmargin=5pt,itemsep=0pt,topsep=0pt,parsep=0pt] 
    \item regression, classification
    \item custom uncertainty quantifiers 
    \end{itemize}
    \end{minipage} \\[6ex]
     BT  &
    \begin{minipage}[t]{\linewidth}
    \begin{itemize}[leftmargin=5pt,itemsep=0pt,topsep=0pt,parsep=0pt] 
    \item  intrinsic (VI-BNN) 
    \end{itemize}
    \end{minipage} & [none] & 
    \begin{minipage}[t]{\linewidth}
    \begin{itemize}[leftmargin=5pt,itemsep=0pt,topsep=0pt,parsep=0pt] 
    \item regression, classification, recurrent  
    \end{itemize}
    \end{minipage} \\[6ex]
    KADF & 
    \begin{minipage}[t]{\linewidth}
    \begin{itemize}[leftmargin=5pt,itemsep=0pt,topsep=0pt,parsep=0pt] 
    \item intrinsic (VI-ADF) 
    \end{itemize}
    \end{minipage} & [none] & 
    \begin{minipage}[t]{\linewidth}
    \begin{itemize}[leftmargin=5pt,itemsep=0pt,topsep=0pt,parsep=0pt] 
    \item regression, classification
    \end{itemize}
    \end{minipage} \\
    \bottomrule
  \end{tabular}
  \end{scriptsize}
  \vspace*{-6pt}
  \caption*{\scriptsize{\textit{Abbreviations}: PL: parametric likelihood, VI-BNN: variational inference-based Bayesian neural networks, GP: Gaussian processes, VI-ADF: assumed density filtering Bayesian neural network (a variant of expectation propagation), MCMC: Markov chain Monte Carlo-based posterior sampling methods, AGC: adversarial group calibration \citep{zhao2020individual}}}
  \bigskip
  
  \begin{scriptsize}
    \begin{tabular}{lp{58mm}p{58mm}}
    \toprule
    \textbf{UE toolkit} &  \textbf{Integration with DL frameworks} & \textbf{Software quality} \\
    \midrule
    TFP & 
    \begin{minipage}[t]{\linewidth}
    \begin{itemize}[leftmargin=5pt,itemsep=0pt,topsep=0pt,parsep=0pt] 
    \item part of tensorflow ecosystem
    \item high integration with keras model API 
    \end{itemize}
    \end{minipage} & 
    \begin{minipage}[t]{\linewidth}
    \begin{itemize}[leftmargin=5pt,itemsep=0pt,topsep=0pt,parsep=0pt] 
    \item high code quality
    \item continuous testing
    \item well-documented 
    \end{itemize}
    \end{minipage} \\[7ex]
    Pyro & 
    \begin{minipage}[t]{\linewidth}
    \begin{itemize}[leftmargin=5pt,itemsep=0pt,topsep=0pt,parsep=0pt] 
    \item pytorch-based implementation
    \item models/inference procedures can be encapsulated as torch modules 
    \end{itemize}
    \end{minipage} & 
    \begin{minipage}[t]{\linewidth}
    \begin{itemize}[leftmargin=5pt,itemsep=0pt,topsep=0pt,parsep=0pt] 
    \item high code quality
    \item continuous testing
    \item well-documented
     \end{itemize}
    \end{minipage} \\[7ex]
    UQ360 & 
    \begin{minipage}[t]{\linewidth}
    \begin{itemize}[leftmargin=5pt,itemsep=0pt,topsep=0pt,parsep=0pt] 
    \item pytorch-based implementation
    \item custom pytorch models can be passed to some modules 
    \end{itemize}
    \end{minipage}
    & 
    \begin{minipage}[t]{\linewidth}
    \begin{itemize}[leftmargin=5pt,itemsep=0pt,topsep=0pt,parsep=0pt] 
    \item simple, easy-to-use sklearn-esque API
    \item well-documented, provides user guidance on uncertainty methods/metrics
    \end{itemize}
    \end{minipage} \\
    \bottomrule
     \end{tabular}
 \end{scriptsize}
 \end{table}
 
\end{document}